\pdfoutput=1

\documentclass[11pt]{article}

\usepackage[]{acl}

\usepackage{times}
\usepackage{latexsym}

\usepackage[T1]{fontenc}

\usepackage[utf8]{inputenc}

\usepackage{microtype}

\usepackage{amssymb}
\usepackage{amsmath}
\usepackage{algorithmic}
\usepackage{algorithm}

\DeclareMathOperator*{\argmax}{arg\,max}

\usepackage[pdftex]{graphicx}

\title{Uncertainty Determines the Adequacy of the Mode and the Tractability of Decoding in Sequence-to-Sequence Models}

\author{Felix Stahlberg$^{1}$ \and Ilia Kulikov$^{2}$\thanks{~~Research done during internship at Google Research, now at Meta AI.} \and Shankar Kumar$^{1}$ \\
\\
    $^{1}$Google Research,  $^{2}$New York University\\
  {\tt \{fstahlberg, shankarkumar\}@google.com}, \\
  {\tt ik1147@nyu.edu}\\
  }

\begin{document}
\maketitle
\begin{abstract}
In many natural language processing (NLP) tasks the same input (e.g.\ source sentence) can have multiple possible outputs (e.g.\ translations). To analyze how this ambiguity (also known as {\em intrinsic uncertainty}) shapes the distribution learned by neural sequence models we measure sentence-level uncertainty by computing the degree of overlap between references in multi-reference test sets from two different NLP tasks: machine translation (MT) and grammatical error correction (GEC). At both the sentence- and the task-level, intrinsic uncertainty has major implications for various aspects of search such as the inductive biases in beam search and the complexity of exact search. In particular, we show that well-known pathologies such as a high number of beam search errors, the inadequacy of the mode, and the drop in system performance with large beam sizes apply to tasks with high level of ambiguity such as MT but not to less uncertain tasks such as GEC. Furthermore, we propose a novel exact $n$-best search algorithm for neural sequence models, and show that intrinsic uncertainty affects model uncertainty as the model tends to overly spread out the probability mass for uncertain tasks and sentences.
\end{abstract}

\section{Introduction}

With the advent of deep learning, many applications of machine learning have converged on a similar set of methods and models. For example, the Transformer \citep{transformer} sequence-to-sequence architecture is ubiquitous in various fields of natural language processing (NLP) such as machine translation (MT), grammatical error correction (GEC), speech recognition \citep{speech-transformer}, etc., and has also been applied successfully to other tasks such as computer vision \citep{vision-transformer}. Recent large pre-trained NLP models such as BERT \citep{devlin-etal-2019-bert}, GPT-3 \citep{gpt3}, T5 \citep{t5}, RoBERTa \citep{roberta}, and XLNet \citep{xlnet} are all based on the Transformer, with relatively minor changes to the architecture itself. 

We show that despite this architectural uniformity the learned distribution over sequences has strikingly different characteristics for different NLP tasks. Inspired by \citet{nmt-uncertainty} we identify {\em intrinsic uncertainty} -- the nature of some NLP tasks to
allow multiple viable outputs for a given input\footnote{This is sometimes referred to as {\em aleatoric} uncertainty in the literature \citep{ml-uncertainty}.} --  to be a major factor that shapes the search space of Transformer models and determines its tractability. In machine translation (MT) -- a task known to have high intrinsic uncertainty \citep{mt-uncertainty,dreyer-marcu-2012-hyter,nmt-uncertainty} -- Transformer models suffer from a high number of beam search errors \citep{stahlberg-byrne-2019-nmt}, an inadequacy of the mode \citep{eikema-aziz-2020-map}, and translation performance degradation with large beam sizes \citep{koehn-knowles-2017-six} (also known as the ``beam search curse''). In contrast, for the correction of writing errors in text (grammatical error correction -- GEC)~\citep{brockett-etal-2006-correcting}, a task with a lower level of uncertainty \citep{bryant-ng-2015-far}, none of these pathologies are evident. This pattern holds even at the sequence-level: input sentences with high uncertainty tend to result in more search errors and a less tractable search space. To study the influence of uncertainty on sequences around the mode, we propose an exact $n$-best search algorithm for neural sequence models. We show that the probability mass covered by the $n$-best candidates differs markedly between certain and uncertain tasks and sentences, which shows that intrinsic uncertainty also affects the spread of probability mass and thus the model uncertainty. We confirm recent work showing that beam search has drawbacks as a decoding scheme for MT. Nevertheless, it is effective for GEC, a problem where modes are adequate, search errors are rare, and the $n$-best lists cover a large fraction of the probability mass.

\section{Measuring Intrinsic Uncertainty}
\label{sec:u}
Intrinsic uncertainty refers to the inherent nature of some NLP tasks to allow for more than one feasible output for a given input. For example, intrinsic uncertainty in MT stems from the fact that there are often several semantically equivalent translations for the same source sentence, or that the translation into a highly inflected language is sometimes under-specified \citep{nmt-uncertainty}. Studies have shown that even for tasks like GEC, annotators do not always agree \citep{tetreault-chodorow-2008-native,rozovskaya-roth-2010-annotating,bryant-ng-2015-far}, but the level of intrinsic uncertainty is arguably lower than for MT because there is a limited number of ways to correct an ungrammatical sentence.

\begin{figure}[t!]
\centering
\small
\includegraphics[scale=1.0]{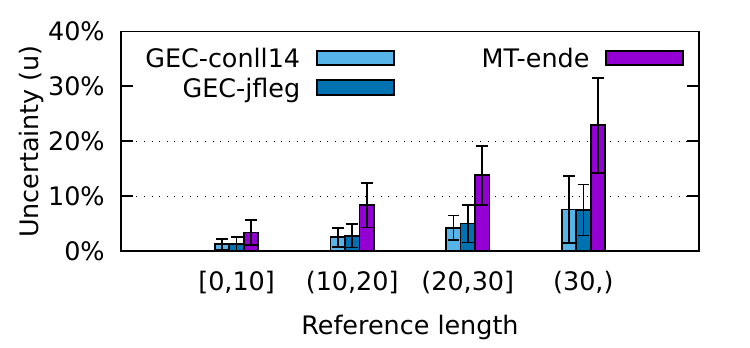}
\caption{Average uncertainty factor $u$ for GEC (blue) and English-German MT (purple) grouped by the sentence length. The error bars show the standard error of the mean (SEM).}
\label{fig:length_lev}
\end{figure}

We propose a simple way to measure sentence-level output uncertainty by making use of multi-reference test sets. For an $n$-way annotated sentence with references $\mathbf{y}_1,...,\mathbf{y}_n$ we define the uncertainty $u$ as the average relative edit distance between two references:
\begin{equation}
\begin{aligned}
u&:=& \underbrace{\frac{1}{\frac{1}{n}\sum_{i=1}^n |\mathbf{y}_i|}}_\text{Avg.\ ref.\ length}\underbrace{\frac{\sum_{i=1}^{n-1} \sum_{j=i+1}^n d_\text{edit}(\mathbf{y}_i, \mathbf{y}_j)}{\frac{n(n-1)}{2}}}_\text{Avg.\ edit distance between refs.}\\
&=& \frac{2}{(n-1)\sum_{i=1}^n |\mathbf{y}_i|} \sum_{i=1}^{n-1} \sum_{j=i+1}^n d_\text{edit}(\mathbf{y}_i, \mathbf{y}_j)
\end{aligned}
\end{equation}
where $d_\text{edit}(\cdot, \cdot)$ denotes the Levenshtein distance. Fig.\ \ref{fig:length_lev} presents this uncertainty score for one MT test set and two GEC test sets. {\em MT-ende} is the official WMT19 English-German test set \citep{barrault-etal-2019-findings} paired with the additional human-annotated  ``\texttt{newstest2019 AR}'' references provided by \citet{freitag-etal-2020-human}.\footnote{The \texttt{AR} references are created from scratch, unlike the other paraphrasing references by \citet{freitag-etal-2020-human}.} {\em GEC-conll14} uses the 10 references published by \citet{bryant-ng-2015-far} for the CoNLL-2014 shared task on GEC \citep{ng-etal-2014-conll}, and {\em GEC-jfleg} is a 4-reference GEC test set that represents ``a broad range of language proficiency levels'' \citep{napoles-etal-2017-jfleg}. Our uncertainty measure reflects our intuition that MT is a significantly more uncertain task than GEC.\footnote{The mean $u$ value differs significantly between GEC and MT in each length bucket ($t$-test $p$-value of less than $0.0001$).} For both tasks the uncertainty increases with the sentence length as longer sentences typically have more feasible mappings than shorter ones.
We use the edit distance rather than task-specific metrics like BLEU \citep{papineni-etal-2002-bleu} or BLEURT \citep{sellam-etal-2020-bleurt} since they
are designed to be robust against uncertainty effects such as reordering or semantically equivalent references, precisely the kinds of effects we aim to capture with $u$. We follow  \citet{bryant-ng-2015-far} by not using inter-annotator agreement statistics like Cohen's $\kappa$ \citep{cohenkappa} since they are more appropriate for the classification into single, well-defined categories.

\section{Mode-seeking Search}

Neural sequence-to-sequence models define a probability distribution $P(\mathbf{y}|\mathbf{x})$ over target sequences $\mathbf{y}$ given a source sequence $\mathbf{x}$:
\begin{equation}
\log P(\mathbf{y}|\mathbf{x}) = \sum_{j=1}^{|\mathbf{y}|} \log P(y_j|y_1^{j-1},\mathbf{x}).
\label{eq:nmt}
\end{equation}
Sequences are typically computed over a subword \citep{sennrich-etal-2016-neural,kudo-richardson-2018-sentencepiece} vocabulary $\mathcal{V}$ and end with a special end-of-sentence symbol $\text{</s>}$:
\begin{equation}
    \mathbf{x},\mathbf{y}\in\{\mathbf{z}\cdot \text{</s>} |\mathbf{z}  \in\mathcal{V}^*\}
\end{equation}
where $\mathcal{V}^*$ is the Kleene closure over $\mathcal{V}$ which includes the empty sequence $\epsilon$. 
Since sequence models are usually trained to maximize the probability of the sequences in the training set, a common strategy to use such a model for inference is to search for the most likely output sequence $\mathbf{y}^*$, also known as the {\em mode} of the model distribution:\footnote{In a Bayesian framework this is often referred to as maximum a posteriori (MAP) inference.}
\begin{equation}
\label{eq:map}
    \mathbf{y}^* = \argmax_\mathbf{y} P(\mathbf{y}|\mathbf{x}).
\end{equation}
Eq.\ \ref{eq:map} is usually approximated using beam search. For analysis purposes, \citet{stahlberg-byrne-2019-nmt} proposed an exact depth-first search (DFS) algorithm that is guaranteed to find the mode.

\section{$N$-best Search}
\label{sec:nbest}

\begin{algorithm}[t!]
\small
\caption{NbestDFS$(n, \mathbf{x},\mathbf{y},p,\gamma, \tilde{Y})$}
\label{alg:nbest-dfs}
\begin{algorithmic}[1]
\REQUIRE{$n$: Search for $n$ global best sequences \\
\hspace{1.5em}$\mathbf{x}$: Source sequence \\
\hspace{1.5em}$\mathbf{y}$: Target prefix (default: $\epsilon$) \\
\hspace{1.5em}$p$: $\log P(\mathbf{y}|\mathbf{x})$ (default: $0.0$) \\
\hspace{1.5em}$\gamma$: Lower bound (default: $-\infty$) \\
\hspace{1.5em}$\tilde{Y}$: Priority queue (default: $\emptyset$)}
\IF{$y_{|\mathbf{y}|}=\text{</s>}$}
  \STATE{$\tilde{Y}.\text{push}((p, \mathbf{y}))$}
  \IF{$|\tilde{Y}| > n$}
    \STATE{$(\gamma,\_) \gets \tilde{Y}.\text{pop}()$}
  \ENDIF
  \RETURN{$(\gamma,\tilde{Y})$}
\ENDIF
\FORALL{$w\in\mathcal{V}$}
  \STATE{$p'\gets p + \log P(w|\mathbf{x},\mathbf{y})$}
  \IF{$p' > \gamma$}
    \STATE{$(\gamma,\tilde{Y})\gets \text{NbestDFS}(n,\mathbf{x}, \mathbf{y}\cdot w, p', \gamma,\tilde{Y})$}
  \ENDIF
\ENDFOR
\RETURN{$(\gamma,\tilde{Y})$}
\end{algorithmic}
\end{algorithm}

In addition to our investigations into the mode we also examine the cumulative probability mass that is covered by the $n$ best hypotheses. If a hypothesis set covers a large fraction of the entire probability mass it approximates the full model distribution well. Approximating the full model distribution is useful for various methods such as minimum risk training \citep{shen-etal-2016-minimum}, reinforcement learning \citep{reinforce,mixer}, minimum Bayes risk decoding \citep{kumar-byrne-2004-minimum,stahlberg-etal-2017-neural,eikema-aziz-2020-map}, etc. \citet{nmt-uncertainty} argued that the fraction of probability mass which is covered by a fixed number of candidates reflects the {\em model} uncertainty on the sequence level. We show that this model uncertainty is in line with our notion of intrinsic uncertainty that we measure with $u$ (Sec.\ \ref{sec:u}). To that end, we propose a generalization of the exact search algorithm of \citet{stahlberg-byrne-2019-nmt} that is able to find the $n$ global best hypotheses rather than the single best one. Similarly to the single-best algorithm, we use the monotonicity of neural sequence model scores:
\begin{equation}
\forall j\in[2,|\mathbf{y}|]: \log P(y_1^{j-1}|\mathbf{x}) > \log P(y_1^j|\mathbf{x}).
\label{eq:monotone}
\end{equation}
\citet{stahlberg-byrne-2019-nmt} keep track of the best complete (i.e.\ ending with the end-of-sentence symbol $\text{</s>}$) hypothesis score during search, and use it to safely prune entire subspaces using Eq.\ \ref{eq:monotone}. In contrast, we keep track of the $n$-th best complete hypothesis score by keeping the $n$ best complete hypotheses in a priority queue. Our exact $n$-best search algorithm is listed in Algorithm \ref{alg:nbest-dfs}. Note that we recover the DFS scheme of \citet{stahlberg-byrne-2019-nmt} with $n=1$.

\section{Experimental Setup}

\begin{table}
\centering
\small
\begin{tabular}{ll}
\hline \textbf{Parameter} & \textbf{Value} \\ \hline
Attention dropout rate & 0.1 \\
Attention layer size & 512 \\
Batch size & 256 \\
Dropout rate & 0.1 \\
Embedding size & 512 \\	
MLP dimension & 2,048 \\	
Number of attention heads & 8 \\
Number of layers & 6 \\
\textbf{Total number of parameters} & \textbf{121M} \\
\hline
\end{tabular}
\caption{\label{tab:trans-hyper} Transformer hyper-parameters.}
\end{table}

\begin{table}
\centering
\small
\begin{tabular}{lll}
\hline \textbf{Language pair} & \multicolumn{2}{c}{\textbf{\#Training sentence pairs}} \\
 & \textbf{Unfiltered} & \textbf{Filtered} \\ \hline
German-English & 39M & 33M \\
Finnish-English & 6.6M & 5.5M \\
Lithuanian-English & 2.3M & 2.0M \\
\hline
\end{tabular}
\caption{\label{tab:mt-train-set} MT training set sizes.}
\end{table}

We trained four Transformer neural machine translation (NMT) models (Table~\ref{tab:trans-hyper}) -- English-German ({\em MT-ende}), German-English ({\em MT-deen}), Finnish-English ({\em MT-fien}), and Lithuanian-English ({\em MT-lten}) -- on the WMT19 \citep{barrault-etal-2019-findings} training sets as provided by TensorFlow Datasets.\footnote{\url{https://www.tensorflow.org/datasets/catalog/wmt19_translate}} We selected these language pairs to experiment with different training set sizes (Table \ref{tab:mt-train-set}). The MT training sets were filtered using language ID and simple length-based heuristics, and split into subwords using joint 32K SentencePiece \citep{kudo-richardson-2018-sentencepiece} models. For training our GEC model we used the hyper-parameters from Table~\ref{tab:trans-hyper} and followed the three-stage training recipe of \citet{stahlberg-kumar-2021-synthetic} using the 32K SentencePiece model from \citet{t5}. All our models were trained until convergence on the development set using the LAMB \citep{lamb} optimizer in JAX \citep{jax} by minimizing cross-entropy without label smoothing. 
Our NMT models are evaluated on the WMT19 test sets \citep{barrault-etal-2019-findings} using SacreBLEU \citep{post-2018-call}. Our GEC model is evaluated on the CoNLL14 \citep[{\em GEC-conll14}]{ng-etal-2014-conll} test set using F$_{0.5}$-scores computed with the M2 scorer \citep{dahlmeier-ng-2012-better} and on the JFLEG test set \citep[{\em GEC-jfleg}]{napoles-etal-2017-jfleg} using GLEU \citep{napoles-etal-2015-ground}.

\section{Results}

\begin{table}
\centering
\small
\begin{tabular}{lllll}
\hline \textbf{System} & \textbf{ende} & \textbf{deen} & \textbf{fien} & \textbf{lten} \\ \hline
\citet{xia-etal-2019-microsoft} & 44.9 & 42.8 & 31.9 & 35.6 \\
Our baselines & 39.6 & 39.7 & 27.7 & 26.9 \\
\hline
\end{tabular}
\caption{\label{tab:comparison-mt} BLEU scores of our NMT baselines and one of the best systems in the WMT19 evaluation campaign -- \texttt{MSRA.MADL} \citep{xia-etal-2019-microsoft}.}
\end{table}

\begin{table}
\centering
\small
\begin{tabular}{lll}
\hline \textbf{System} & \textbf{conll14 (F$_{0.5}$)} & \textbf{jfleg (GLEU)} \\ \hline
\citet{lichtarge-etal-2020-data} & 66.8 & 64.9 \\
\citet{rothe-etal-2021-simple} & 68.9 & - \\
Our baseline & 60.0 & 62.1 \\
\hline
\end{tabular}
\caption{\label{tab:comparison-gec} Comparison of our GEC baseline with the best results reported in the literature.}
\end{table}

In this work our focus is to analyze the impact of intrinsic uncertainty on search. Thus we keep our setup simple, reproducible, and computationally economical rather than obtain new state-of-the-art results. Nevertheless, Tables \ref{tab:comparison-mt} and \ref{tab:comparison-gec} show that our baselines are not unreasonably far off from the best results in the literature given that the systems we compare with are often highly engineered and use many more parameters. \citet{xia-etal-2019-microsoft} used various techniques like back-translation, ensembling, dual learning, MASS pre-training, architecture search, larger models, etc.\ to improve their systems, and \citet{rothe-etal-2021-simple} used a 11B parameters T5 \citep{t5} model.

\subsection{Finding the Most Likely Hypothesis}
\label{sec:mode-results}

Even though alternative decision rules like MBR have recently received some attention in the NMT literature \citep{eikema-aziz-2020-map,muller-sennrich-2021-understanding}, mode-seeking decoding schemes such as beam search or Nucleus sampling \citep{nucleus} are by far the most common choices. In this section we explore how uncertainty changes the mode and the ability of beam search to find it.

\begin{figure}[t!]
\centering
\small
\includegraphics[scale=1.0]{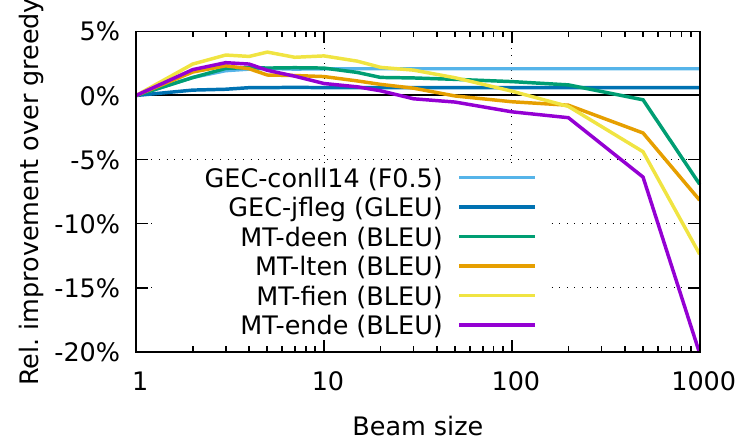}
\caption{Relative beam search improvements over greedy search. MT quality degrades with large beam sizes, but GEC saturates after a beam size of 10.}
\label{fig:beam_improvement_over_greedy}
\end{figure}

A well-known pathology of NMT models is the ``beam search curse'' \citep{koehn-knowles-2017-six}: Increasing the beam size improves the predictive log-probabilities of the hypotheses, but it leads to worse translation quality due to the NMT model error of preferring short translations.
We replicate this result in Fig.\ \ref{fig:beam_improvement_over_greedy}: BLEU scores for MT initially improve over greedy search at smaller beam sizes but after reaching a peak at beam size of 4, we observe a dramatic drop in BLEU. The trajectory of the blue curves (GEC) is markedly different: the performance does {\em not} drop for large beams but saturates instead. The beam search curse affects tasks with high intrinsic uncertainty like MT but spares more certain tasks like GEC although both tasks use the same neural Transformer architecture.

\begin{figure}[t!]
\centering
\small
\includegraphics[scale=1.0]{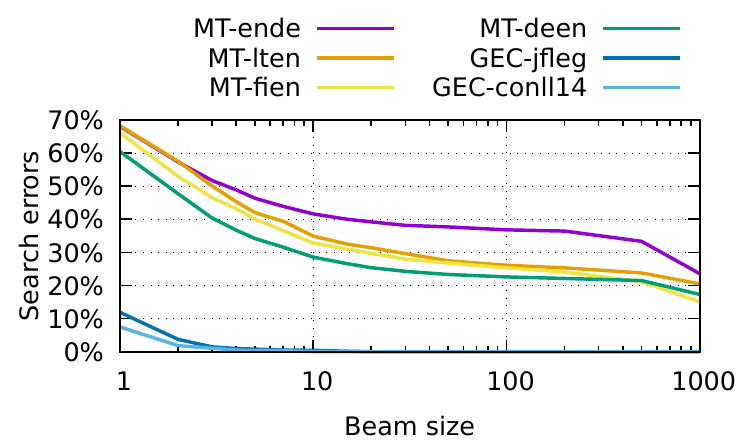}
\caption{Number of beam search errors.}
\label{fig:beam_search_errors}
\end{figure}

To determine why the beam size affects NMT and GEC so differently we ran the exact decoding algorithm of \citet{stahlberg-byrne-2019-nmt} to find the global best hypotheses and counted search errors, i.e.\ the number of sentences in the test set for which beam search does not find the global best sequence. Our results confirm the findings of \citet{stahlberg-byrne-2019-nmt} that increasing the beam sizes leads to fewer NMT search errors (Fig.\ \ref{fig:beam_search_errors}). Among our MT language pairs, English-German (MT-ende) suffers the most from the beam search curse and the proportion of search errors in the test set. This is possibly because translation from English to German typically results in a longer sequence and thus more uncertainty. GEC differs significantly from NMT in the total number of search errors. For MT, even with a very large beam size of 500, beam search does not find the mode for more than 20\% of the sentences in any language pair. In contrast for GEC, we do not observe any search errors for beam sizes larger than 10. This suggests that task uncertainty determines the tractability of the search space and particularly the search for the mode.

Uncertainty also determines the computational costs of exact search. To abstract away from hardware and implementation details, we measure the time complexity of exact search by counting the number of explored states, i.e.\ the number of forward passes through the model, which is identical to the number of recursive calls of Algorithm \ref{alg:nbest-dfs}.\footnote{For example, the number of explored states in standard beam search is the beam size times the target sequence length.} 

\begin{figure}[t!]
\centering
\small
\includegraphics[scale=1.0]{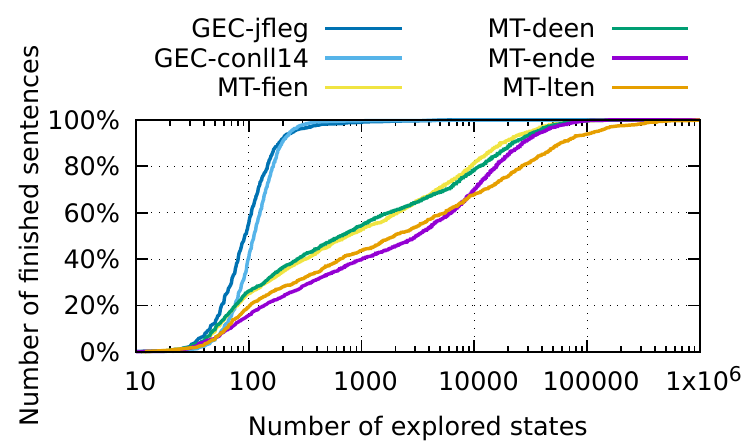}
\caption{Number of states exact search needs to explore in order to find and verify the mode.}
\label{fig:states_sentences_n1}
\end{figure}

\begin{figure*}[t!]
\centering
\small
\begin{tabular}{@{\hspace{0em}}c@{\hspace{0em}}c@{\hspace{0em}}}
\multicolumn{2}{c}{\includegraphics[scale=0.16]{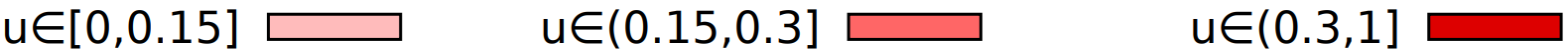}} \\
\includegraphics[scale=1.0]{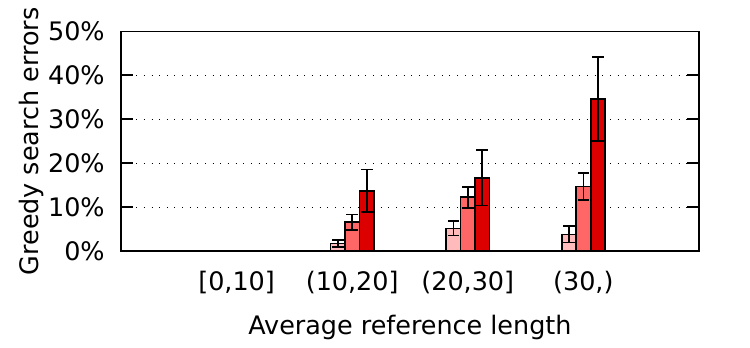} & \includegraphics[scale=1.0]{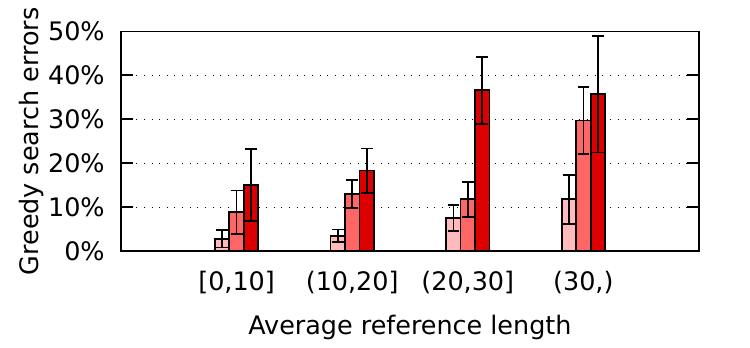} \\
(a) Greedy search errors (GEC-conll14) & (b) Greedy search errors (GEC-jfleg) \\
\includegraphics[scale=1.0]{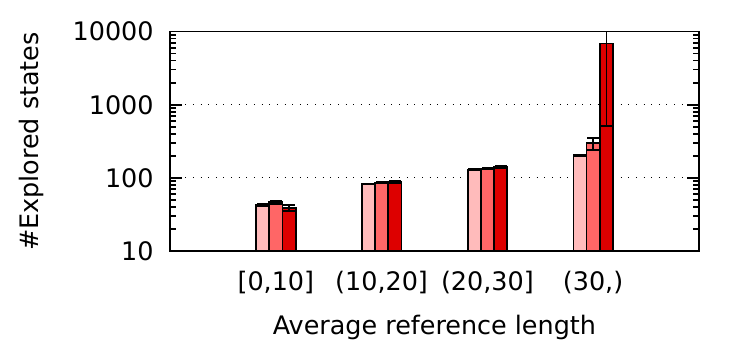} & \includegraphics[scale=1.0]{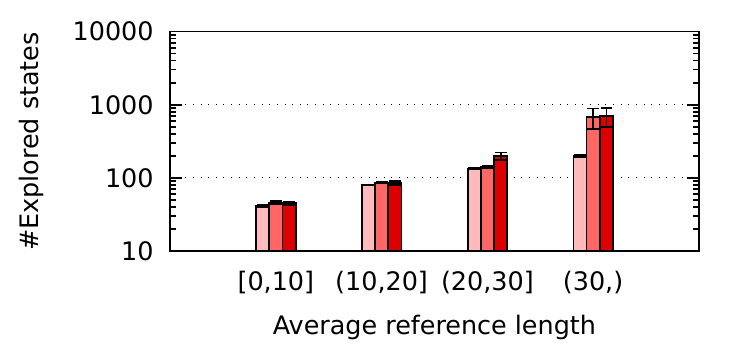} \\
(c) Number of explored DFS states (GEC-conll14) & (d) Number of explored DFS states (GEC-jfleg) \\
\end{tabular}
\caption{The impact of sentence length and uncertainty $u$ on the number of greedy search errors and the number of explored states by exact search for GEC. The error bars show the SEM.}
\label{fig:length_lev_greedy_errors_states_gec}
\end{figure*}

\begin{figure}[t!]
\centering
\small
\begin{tabular}{@{\hspace{0em}}c@{\hspace{0em}}}
\includegraphics[scale=0.16]{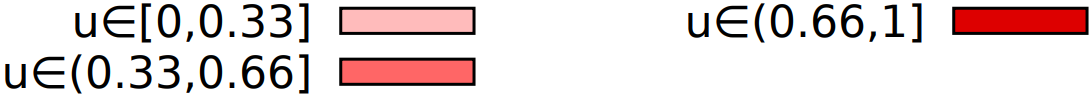} \\
\includegraphics[scale=1.0]{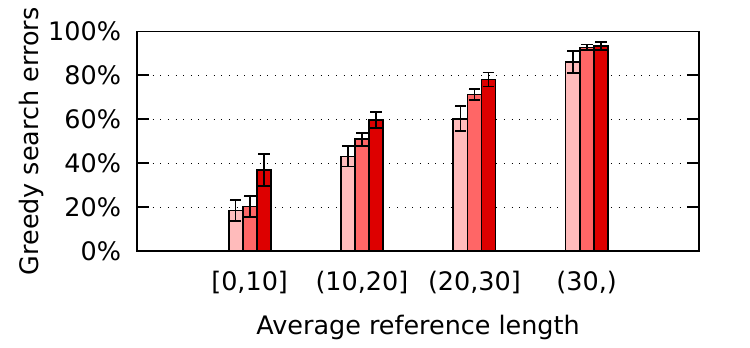}  \\
(a) Greedy search errors (MT-ende) \\
\includegraphics[scale=1.0]{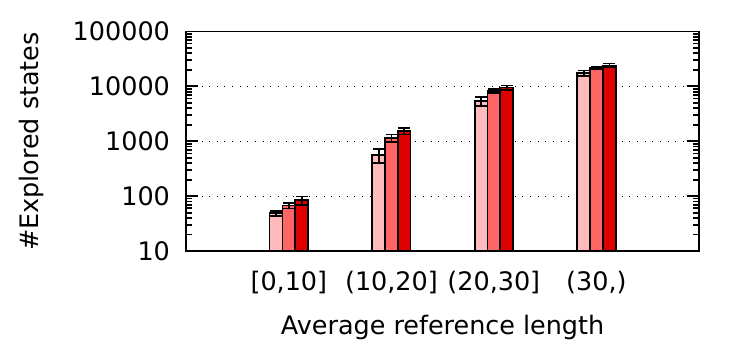}  \\
(b) Number of explored DFS states (MT-ende) \\
\end{tabular}
\caption{The impact of sentence length and uncertainty $u$ on the number of greedy search errors and the number of explored states by exact search for MT. The error bars show the SEM.}
\label{fig:length_lev_greedy_errors_states_mt}
\end{figure}

Fig.\ \ref{fig:states_sentences_n1} plots the fraction of sentences in the test set for which the exact search explores a certain maximum number of states to terminate. For example, exact search returned the mode for around 50\% of the MT sentences after exploring no more than 1000 states. With the same computational budget, however, it was able to find the mode for nearly 100\% of the GEC sentences (blue curves). For some of the MT sentences, exact search needed to explore around 100K states, or even more in the case of Lithuanian-English (orange curve).

\paragraph{Sentence-level uncertainty}

In the previous paragraph we showed that MT, a task with high intrinsic uncertainty, suffers from more beam search errors and a less tractable search space than GEC, a task with relatively low intrinsic uncertainty. Figs. \ref{fig:length_lev_greedy_errors_states_gec} and \ref{fig:length_lev_greedy_errors_states_mt} demonstrate that this pattern is not only present at the task-level but also at the sentence-level. First, the bar charts show that there is a general trend towards more search errors and more explored states for longer sentences. Longer input sentences often result in higher entropy distributions (i.e.\ more uncertainty) since there are usually more ways to map a long sentence than a short one. We also see a pattern within each group, i.e.\ within a reference length interval, that shows that sentences with higher uncertainty $u$ result in more search errors and a longer exact search runtime even when compared to other sentences with similar lengths. Table \ref{tab:correlation} lists the test set level correlation coefficients.

\subsection{The Spread of Probability Mass}

\begin{figure}[t!]
\centering
\small
\includegraphics[scale=1.0]{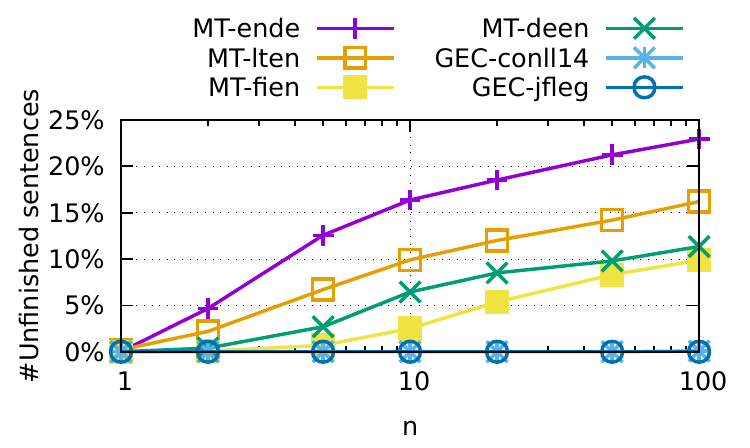}
\caption{Number of sentences for which exact $n$-best search did not terminate before 1M explored states.}
\label{fig:dfs_maxedout}
\end{figure}

\begin{figure}[t!]
\centering
\small
\includegraphics[scale=1.0]{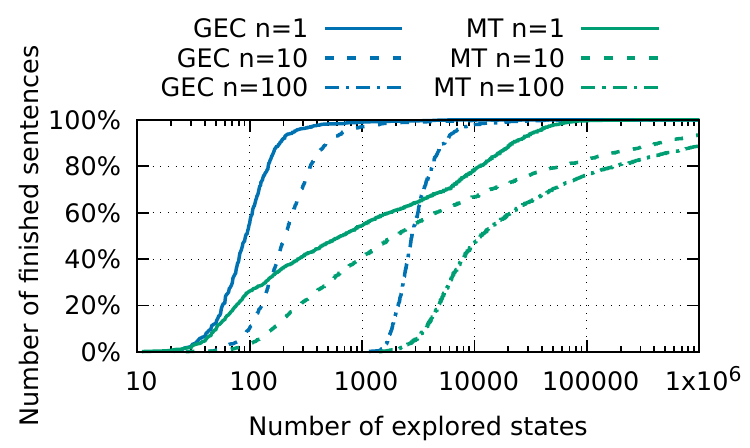}
\caption{Number of states exact $n$-best search needs to explore in order to terminate for GEC-jfleg and MT-deen.}
\label{fig:states_sentences_n1_10_100}
\end{figure}

We argued in Sec.\ \ref{sec:nbest} that the ability to approximate the entire search space with a fixed set of candidates can be useful in training \citep{shen-etal-2016-minimum,reinforce,mixer} and decoding \citep{kumar-byrne-2004-minimum,eikema-aziz-2020-map}, and proposed an exact $n$-best search algorithm. However, finding the
exact $n$-best hypotheses is computationally much more expensive than finding the single-best hypothesis (mode). Therefore, to keep the runtime under control, we stopped $n$-best decoding after 1M explored states. Fig.\ \ref{fig:dfs_maxedout} shows that the 1M threshold is not reached for $n=1$ for any sentence: it was always possible to find and verify the mode. We can guarantee that the $n=100$ best candidates returned by our algorithm are indeed the global best ones for around 90\% of the MT-deen sentences (right end of the green curve in Fig.\ \ref{fig:dfs_maxedout}). 
The blue curves in Fig.\ \ref{fig:dfs_maxedout} suggest that as before the GEC search space is much more tractable given that our exact $n$-best search algorithm was able to find the 100 global best hypotheses for all GEC sentences before reaching 1M explored states. Indeed, Fig.\ \ref{fig:states_sentences_n1_10_100} shows that exact 100-best search terminated with fewer than 10K explored states for almost all GEC sentences while the pruning criterion in Eq.\ \ref{eq:monotone} is much less effective for the NMT search space (green curves in Fig.\ \ref{fig:states_sentences_n1_10_100}).

\begin{figure}[t!]
\centering
\small
\includegraphics[scale=1.0]{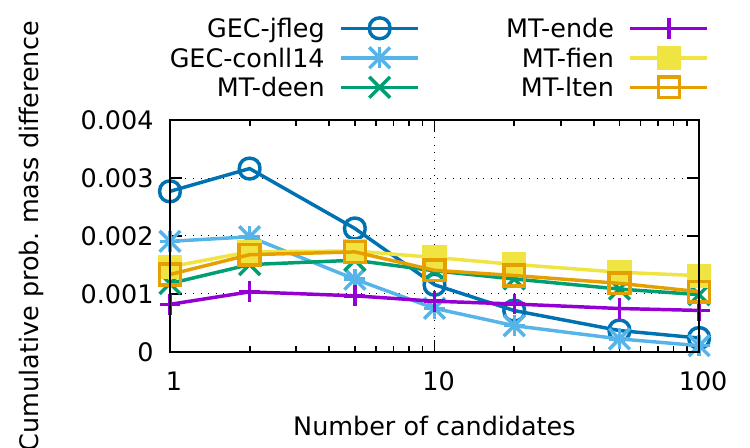}
\caption{Difference in cumulative probability mass between the global $n$ best hypothesis set returned by exact $n$-best search and the $n$-best list returned by beam search with different beam sizes.}
\label{fig:prob_mass_diff}
\end{figure}

The cumulative probability mass of the set returned by exact $n$-best search is an upper bound for the cumulative probability mass of any hypothesis set with a cardinality of $n$. Despite the high number of search errors (Fig.\ \ref{fig:beam_search_errors}), the probability mass covered by the $n$-best beam search hypotheses is very close to this upper bound. Fig.\ \ref{fig:prob_mass_diff} shows that for $n=100$ that difference is less than 0.001 for all setups except MT-fien. Since the difference in probability mass is negligible we ran our subsequent investigations of probability mass with beam search instead of exact search to save computational costs.

\begin{figure}[t!]
\centering
\small
\includegraphics[scale=1.0]{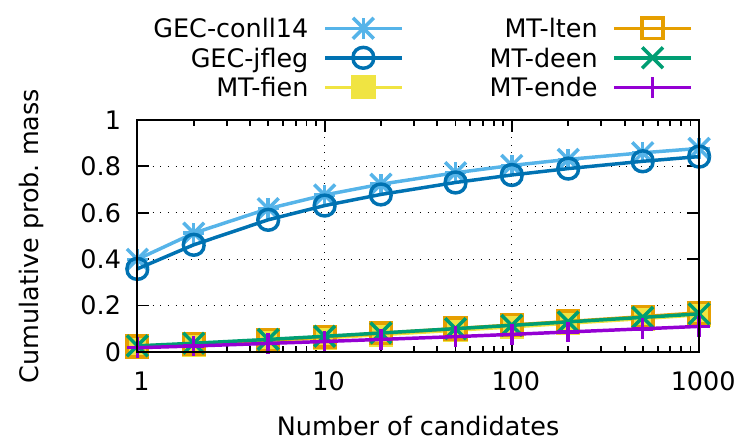}
\caption{Average probability mass covered by the $n$-best list from beam search for beam sizes between 1 and 1000.}
\label{fig:beam_avg_prob_mass}
\end{figure}

\begin{figure}[t!]
\centering
\small
\includegraphics[scale=1.0]{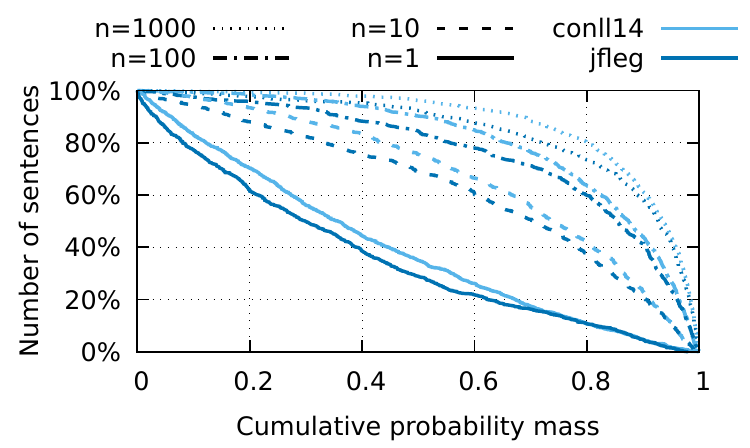}
\caption{The number of sentences for which the total probability mass contained in a beam search $n$-best list with beam sizes of 1, 10, 100, 1000 is a certain fraction of the total probability mass (GEC).}
\label{fig:beam_prob_mass_sentences_gec}
\end{figure}

\begin{figure}[t!]
\centering
\small
\includegraphics[scale=1.0]{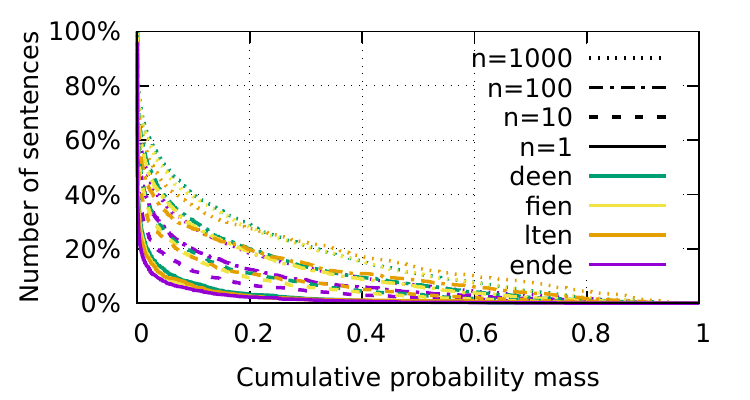}
\caption{The number of sentences for which the total probability mass contained in a beam search $n$-best list with beam sizes of 1, 10, 100, 1000 is a certain fraction of the total probability mass (MT).}
\label{fig:beam_prob_mass_sentences_mt}
\end{figure}

Fig.\ \ref{fig:beam_avg_prob_mass} visualizes the difference between NMT and GEC in terms of the probability mass covered by the beam search hypotheses.  We confirm the finding of \citet{nmt-uncertainty,eikema-aziz-2020-map} that the NMT distribution is rather flat: even 1000 MT candidates cover only 20\% of the probability mass on average. In GEC, however, the model assigns twice as much probability (40\%) to the {\em single} best hypothesis on average (left end of the blue curves in Fig.\ \ref{fig:beam_avg_prob_mass}). Fig.\ \ref{fig:beam_prob_mass_sentences_gec} provides even more insight: A beam size of 1000 covers 40\% of the probability mass for nearly all sentences in the GEC test sets. Even more practical beam sizes of 10 cover more than half of the probability mass for around 75\% of the GEC-conll14 sentences. The same plot looks very different for MT (Fig.\ \ref{fig:beam_prob_mass_sentences_mt}): Covering half the probability mass is only possible for a tiny fraction of the MT sentences.

\begin{figure*}[t!]
\centering
\small
\begin{tabular}{@{\hspace{0em}}c@{\hspace{0em}}c@{\hspace{0em}}}
\multicolumn{2}{c}{\includegraphics[scale=0.16]{key_length_lev_gec.png}} \\
\includegraphics[scale=1.0]{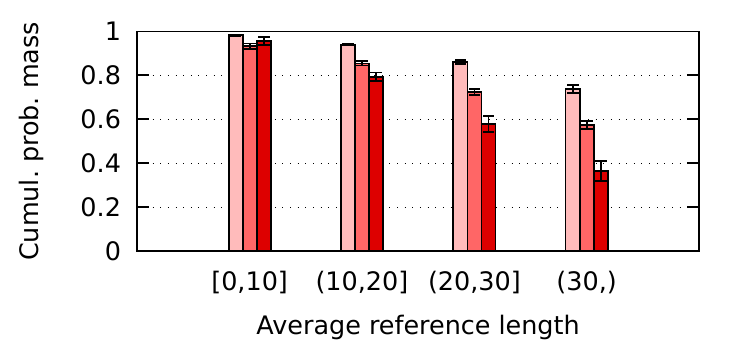} & \includegraphics[scale=1.0]{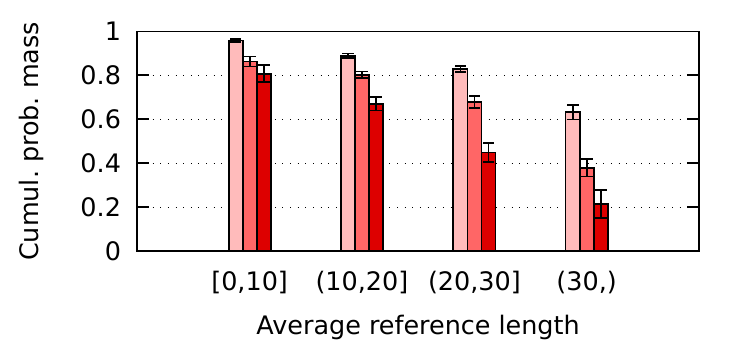} \\
(a) GEC-conll14 & (b) GEC-jfleg \\
\end{tabular}
\caption{The impact of sentence length and uncertainty $u$ on the cumulative probability mass of the 100-best list from beam search for GEC. The error bars show the SEM.}
\label{fig:length_lev_beam100_probsum_gec}
\end{figure*}

\begin{figure}[t!]
\centering
\small
\begin{tabular}{@{\hspace{0em}}c@{\hspace{0em}}}
\includegraphics[scale=0.16]{key_length_lev_mt_singlecol.png} \\
\includegraphics[scale=1.0]{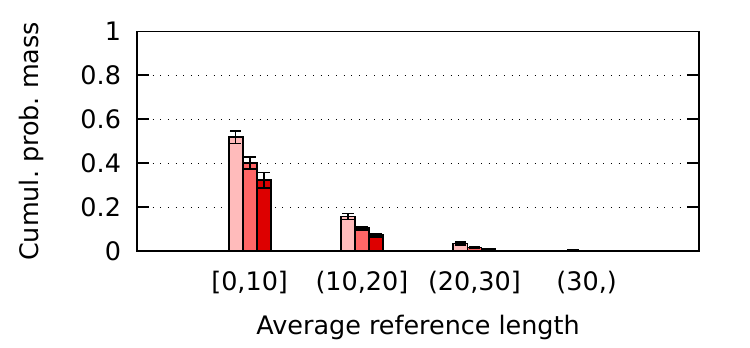}  \\
\end{tabular}
\caption{The impact of sentence length and uncertainty $u$ on the cumulative probability mass of the 100-best list from beam search for MT. The error bars show the SEM.}
\label{fig:length_lev_beam100_probsum_mt}
\end{figure}

\begin{table}[t!]
\centering
\small
\begin{tabular}{lccc}
\hline \textbf{$\rho$ between $u$ and\dots} & \multicolumn{2}{c}{\textbf{GEC}} &  \textbf{MT} \\
 & \textbf{conll14} & \textbf{jfleg} & \textbf{ende} \\ \hline
Greedy search errors & \ 0.18 & \ 0.19 & \ 0.24 \\
\#Explored DFS states & \ 0.20 & \ 0.18 & \ 0.19 \\
Cumul.\ prob.\ mass & -0.23 & -0.51 & -0.53 \\
\hline
\end{tabular}
\caption{\label{tab:correlation} Spearman's rank correlation coefficient $\rho$ between the uncertainty $u$ and the number of greedy search errors, the number of explored DFS states, and the 100-best cumulative probability mass. All correlations are significant with a $p$-value of less than 0.00001.}
\end{table}

\paragraph{Sentence-level uncertainty}

In Sec.\ \ref{sec:mode-results} we reported that the effects caused by intrinsic uncertainty on the ability to find the mode are visible at both the task- and the sentence-levels. Similarly, we can track down our observations about how uncertainty determines the probability mass of $n$-best lists at the sentence level. Fig.\ \ref{fig:length_lev_beam100_probsum_gec} shows that the cumulative probability mass in the $n$-best list decreases for longer sentences as the mappings of long sentences are more uncertain. Again, the trend within a group in Fig.\  \ref{fig:length_lev_beam100_probsum_gec} suggests that even among sentences with similar lengths, $n$-best lists for uncertain sentences (higher $u$) accumulate less probability mass. We make analogous observations for NMT (Fig.\ \ref{fig:length_lev_beam100_probsum_mt}), although the total $n$-best probability mass is much smaller than for GEC.

\section{Related Work}
\label{sec:related-work}

Ambiguity is one of the core challenges in MT, a fact that is supported (inter alia) by the long history of designing evaluation metrics that are robust against it \citep{papineni-etal-2002-bleu,banerjee-lavie-2005-meteor,sellam-etal-2020-bleurt}. In this work we examine the impact of ambiguity on the NMT search space, and show how it is related to various well-known issues of NMT models like the beam search curse \citep{koehn-knowles-2017-six}, a pathology that has also been linked to the local normalization in sequence models \citep{sountsov-sarawagi-2016-length,murray-chiang-2018-correcting} or poor model calibration \citep{nmt-calibration}.

Our work is heavily inspired by \citet{nmt-uncertainty} who analyzed different kinds of uncertainty in NMT. In particular, they found that NMT spreads out the probability mass over a large number of candidates, and connected the beam search curse with uncertainty. We confirm their results and extend their line of research along the following directions:
We introduce a measure for uncertainty in multi-reference test sets, and show that the negative effects of uncertainty are visible even on the sentence level. 
Second, we propose an exact $n$-best search algorithm and demonstrate how it can be used to analyze the spread of probability mass. Third, we focus not only on MT but also on GEC.

\citet{stahlberg-byrne-2019-nmt} showed that beam search errors often obscure the length deficiency of the NMT modes, and reducing search errors by using large beams exposes this model error. In this work, we found that these mechanics are limited to NMT: GEC does not suffer from the beam search curse since search errors are rare and modes are not too short.   
\citet{eikema-aziz-2020-map} suggested that picking a hypothesis based solely on probability is erratic because NMT spreads out the probability mass over a large set of hypotheses with similar probabilities. Therefore, alternative approaches that in addition to the probabilities incorporate MT-specific metrics such as BLEU \citep{papineni-etal-2002-bleu} or BLEURT \citep{sellam-etal-2020-bleurt} have recently been in focus of research, including minimum Bayes risk decoding \citep{eikema-aziz-2020-map,eikema2021sampling,muller-sennrich-2021-understanding}, Monte-Carlo tree search \citep{leblond2021machine}, and energy-based \citep{bhattacharyya-etal-2021-energy} or discriminatively trained \citep{lee-etal-2021-discriminative} rerankers. Our work on how uncertainty determines the spread of probability mass is relevant to those approaches.


\section{Conclusion}

We identified a major culprit behind various inference-related issues in sequence-to-sequence models such as the intractability of the search space, degenerate large beam or exact search outputs and the large spread in probability mass over the output space. This factor is intrinsic uncertainty -- the existence of multiple ways to correctly map an input sequence. We measured the intrinsic uncertainty of input sentences as the degree of agreement between multiple references and showed that ambiguous sentences typically result in a higher number of beam search errors and an exceedingly flat output distribution. We also find that known NMT pathologies such as the beam search curse or inadequate modes do not extend to less ambiguous tasks like GEC despite using the same neural architecture.


\bibliography{anthology,custom}
\bibliographystyle{acl_natbib}






\end{document}